# Face Attribute Invertion

X G TU[1], Y LUO[1], H S ZHANG[1], W J AI[1], Z MA[1], and M XIE[1]

[1] School of Information and Communication Engineering, University of Electronic Science and Technology of China. Chengdu, China.

xguangtu@outlook.com, 15882421574

**Abstract.** Manipulating human facial images between two domains is an important and interesting problem. Most of the existing methods address this issue by applying two generators or one generator with extra conditional inputs. In this paper, we proposed a novel self-perception method based on GANs for automatical face attribute inverse. The proposed method takes face images as inputs and employs only one single generator without being conditioned on other inputs. Profiting from the multi-loss strategy and modified U-net structure, our model is quite stable in training and capable of preserving finer details of the original face images.

## 1. Introduction

Face attribute manipulation is a task to edit face attributes presented in an image, e.g. facial expression, emotion, age and so on. Compared with face style transfer task [1][19], it is more challenging due to the requirement of only modifying some specific regions in the source image while keeping other regions unchanged. Thanks to the generative adversarial networks (GANs)[2][16][17][18], this task has experienced significant improvement, with the quality of synthetic images highly improved.

Many methods utilize original images as inputs and try to reconstruct the finer details of the original images when modifying the interest areas of the input images. IcGAN[3] leverages conditional code and a latent representation inferred from a given image by a pre-trained encoder to generate desired image. Nevertheless, the quality of the reconstruction from the latent representation is not well-pleasing as there is not an explicit mechanism for inverse mapping of an input image to the latent vector which is necessary for image reconstruction. Recent works [6][7][20] have witnessed great success in the task of image-to-image translation which takes original images as input and outputs the transformed ones without explicit embedding. For instance, lsola Phillp et al.[4] have proposed the U-net (see Fig. 1(b)), which adds skip connections between the symmetrical layers of the encoder and decoder, to facilitate the reusage of information during backpropagation. Inspired by their architecture, we proposed a new approach based on the structure of U-net for the purpose of face attribute manipulation. The proposed approach could automatically perceive the "mode" of the facial attributes present in the input image and learn to translate it to its inverse pattern. To ensure image consistency and preserve key information between the input and output face images, cycle consistency loss (from CycleGAN[5]) was applied to the generator. However, different from CycleGAN where two separated generators are used for domain "A→B" and "B→A" transformations respectively, only one generator network was employed in our method.

To guide the generated images towards certain patterns, methods like[6][7][8] have been proposed with an additional classifier embedded in the discriminator of GANs. For example, A.Odenaetal.[6]

propose the AC-GAN, using the auxiliary classification loss to ensure the generated images globally coherent with the input data. In the work [7], L. Zhang et al. modify the discriminator in AC-GAN to produce paintings with different styles. Similarly, StarGAN[8] introduces an auxiliary classifier that uses an additional discriminator to control multiple domains.

Following their works, we extended the model to a multi-task setting, in which the discriminator differentiates face images between fake and real, and meanwhile handles classification task with respect to facial attributes by minimizing the classification error related to the known attribute labels. Generally, our contributions are summarized as follows.

- A GAN structure with only one generator is proposed for face attribute manipulation without any conditional inputs offered.
- A modified U-net is proposed to ensure attribute specific areas easier to be manipulated.
- A new quantitative analysis metric is proposed to evaluate the generated images' quality.

## 2. The Proposed Method

Our method aims to train a generator $G$ which can perceive the "mode" of the facial attributes present in the input image and translate it to its inverse pattern, the architecture is presented in figure 1(a). Our model addresses each of the facial attributes independently. The face generator $G$ is trained to generate a photo-realistic face image from the observed face image $x$: $X_{fake} = G(x)$, and the auxiliary classifier discriminator $D$ imposes a distribution constraint over the source images, along with a probability distribution over the attribute labels, $P(S|X) = D_s(x), P(C|X) = D_c(x)$.

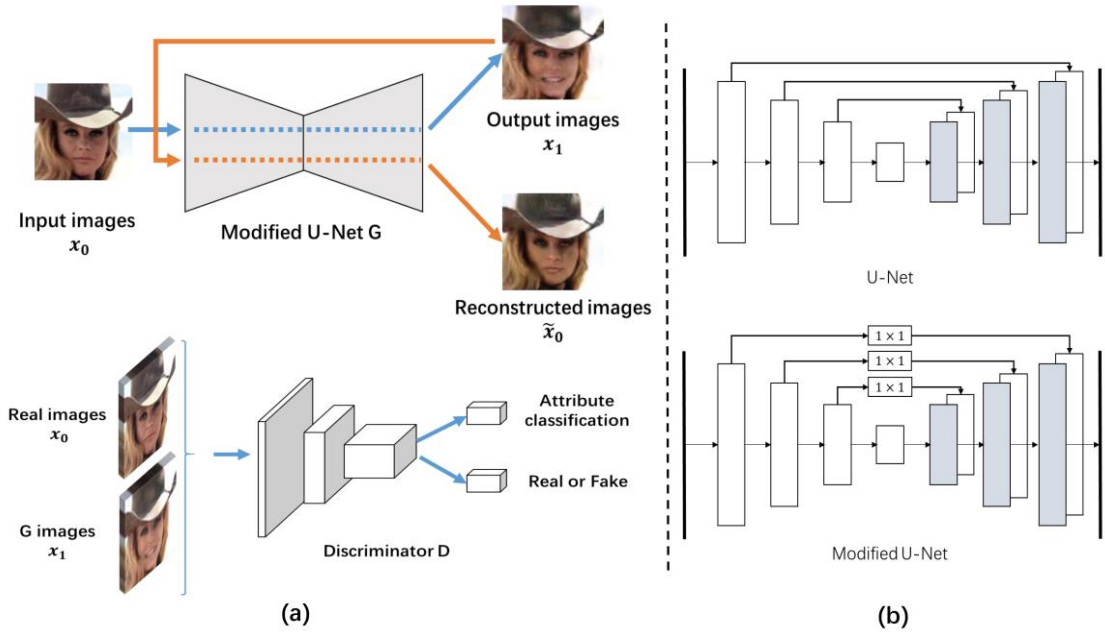

**Figure 1.** (a): The blue line in the upper framework indicates taking input image $x_0$ as input and output attribute-inverse image $x_1$, while the orange line represents taking $x_1$ as input and generate the reconstruction $\tilde{x}_0$. The nether structure is the multi-task discriminator network $D$. (b): The architecture of the original U-Net (upper) and our modified one(nether).

### 2.1. Objective

During training, the process is supervised by an auxiliary classifier discriminator $D$. In the min-max training process, $G$ generates images that cannot be differentiated from real ones with the face attributes opposite to input images while *D* is trained in a joint multi-task fashion that combines the task of distinguishing real/generated images and the task of attributes classification.

*2.1.1 Adversarial loss.* Adversarial loss $L_{adv}$ was leveraged to make the generated images more realistic, it is defined as

$$L_{adv} = \mathrm{E}_x[logD_s(x)] + \mathrm{E}_x[1 - logD_s(G(x))], \quad (1)$$

where generator *G* maps the given face image *x* to the output *G(x)* to fool *D*, and *D* tries to differentiate the generated image *G(x)* from the real one by the terms $D_s(x)$. So *D* is trained to maximize $L_{adv}$ while *G* is trained to minimize it. The parameters of *G* and *D* are iteratively trained and updated.

*2.1.2 Attribute classification loss.* In our method, *G* is trained to generate modified images where their attributes are opposite to the original ones. So we extended an attribute classification task on the top of *D* by the classification loss $L_{cls}$ to force the generator to produce attribute-inverse images. The losses from original images and fake images are respectively defined as

$$L_{cls_{real}} = \mathrm{E}_{x,c}[-logD_{cls}(c|x)] \quad (2)$$

$$L_{cls_{fake}} = \mathrm{E}_{x,c'}[-logD_{cls}(c'|G(x))], \quad (3)$$

where the term $D_{cls}(x)$ represents a probability distribution over face attributes. $L_{cls_{real}}$ is calculated from the real face images, it is used to optimize *D*, and $L_{cls_{fake}}$ which is calculated from the outputs, it is used to optimize *G*. *c* and *c'* denote original attribute label and inverse attribute label, respectively.

*2.1.3 Forward-backward consistency loss.* By minimizing attribute classification loss and the adversarial loss, *G* is capable of producing photorealistic face images with inverse attributes that are expected to be changed. To further preserve the attribute-irrelevant details of input images in the translated images, we employed a reconstruction loss $L_{rec}$ and a feature matching loss $L_{fm}$ as forward backward consistency loss to measure the differences between real image *x* and reconstructed image *G(G(x))* on pixel and features level respectively. The reconstruction loss $L_{rec}$ is computed on pixel level:

$$L_{rec} = \mathrm{E}_x[\|x - G(G(x))\|_1], \quad (4)$$

where the cost function is based on L1 norm which encourages less blurring than L2 sense[4]. The feature matching loss $L_{fm}$[9] is based on the discriminator *D*, it is used to stabilize the training process. In details, we extract the features from multiple layers of the discriminator, learning to match these representations from the real image *x* and the reconstructed image *G(G(x))*. It is defined as

$$L_{fm} = \mathrm{E}_x \sum_{i=1}^{T} \frac{1}{N_i}[\|D_i(x) - D_i(G(G(x)))\|_1], \quad (5)$$

where the term $D_i(x)$ represents the *i*-th layer feature representation in the discriminator *D*, $N_i$ denotes the number of pixels in each layer and *T* is the total number of layers.

The final objective function of *G* and *D* are defined by

$$L_G = L_{adv} + \lambda_1 L_{cls_{fake}} + \lambda_2 L_{rec} + \lambda_3 L_{fm} \quad (6)$$

$$L_D = -L_{adv} + \lambda_4 L_{cls_{real}}, \quad (7)$$

Where $\lambda_i$, *i* = 1,2,3,4 denote hyper-parameters that balance the weights of different loss functions.

*2.2. Network Details*

As we know, the hidden layers contained in CNN allow the network learn representations hierarchically from lower to higher level[10]. Representations from lower layers respond to general features in the images, such as edge, color, texture, etc., while the representations from middle to higher layers are progressively class-specific and contain information from partial to complete regarding the objects. In the case of face attribute manipulation, we would like to only modify the interest areas (e.g., month, hair and glasses) and meanwhile keep other regions untouched. While the original U-net tends to remain all details of original images in the generated images due to full skip connections between all corresponding layers, which makes the modification of attribute-relevant area not very obvious. To address this issue, we added 1×1 filters between corresponding layers of the encoder and decoder to reduce redundant information. Meanwhile, we set different numbers of 1×1 filters for different skip connections to control the shared contents. In experiments, the number of 1×1 filters for high-level skip connections was set to half of the number of input feature maps, while one quarter for low-level ones. Hence, the modified U-net could share more class-specific contents and object information which makes the attribute-relevant areas easier to be manipulated and some high-level information like identity more likely to be preserved. The structures of the original U-net and our modification are illustrated in figure 1(b). An attribute classifier was added on top of discriminator $D$ to give both a probability distribution over the source images and a probability distribution over the attribute labels (see figure 1(a)). In this way, the discriminator can produce images not only are indistinguishable from real images but also possess the inverse attributes compared with original images.

## 3. Experiment

*3.1. Experimental Settings*

Our model was trained on the CelebFaces Attributes (CelebA) dataset[11] which contains more than 200, 000 celebrity images, annotated with 40 binary attributes, such as eyeglasses, male, pale skin, etc.

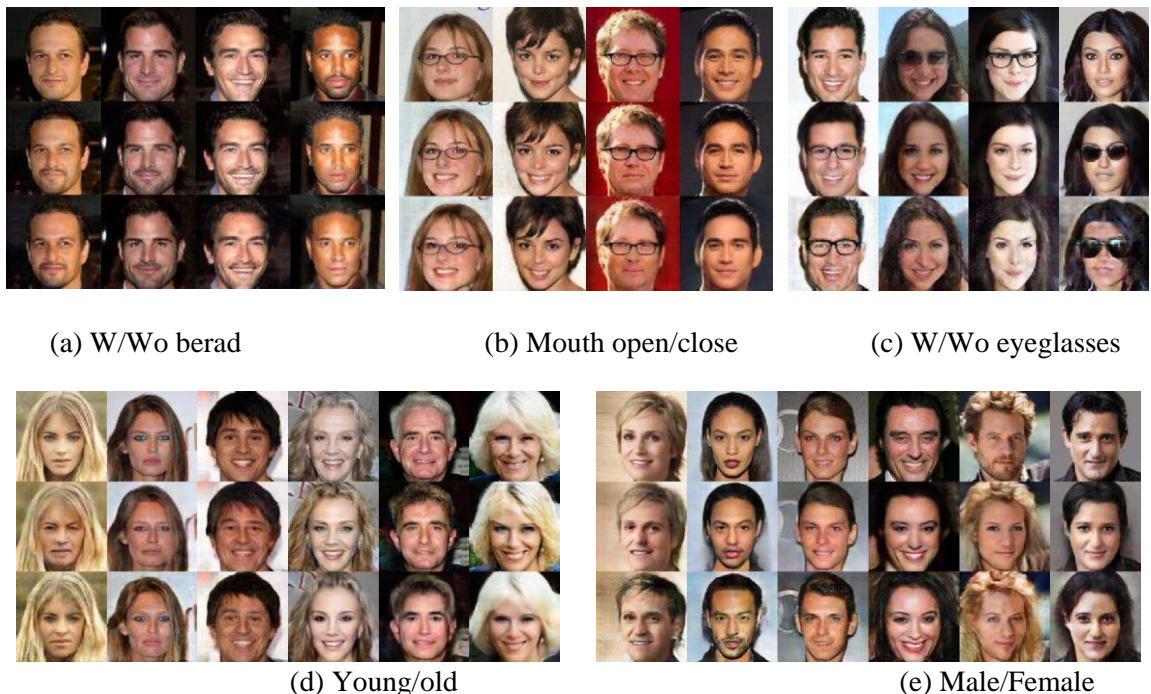

(a) W/Wo berad  (b) Mouth open/close  (c) W/Wo eyeglasses

(d) Young/old  (e) Male/Female

**Figure 2.** Face attribute manipulation on the CelebA dataset. For each sub-figure, the first row shows the original face images. The second and the third row are the manipulated images using the proposed method and CycleGAN, respectively.

The center part of the aligned images are cropped and scaled to 128×128. As the attribute labels are highly biased in the CelebA dataset, we employed a simple oversampling strategy[12] to relieve the impact of imbalances in the training process. The hyper-parameters $\lambda_1$, $\lambda_3$, $\lambda_4$ were set to 1, while $\lambda_2$ was set to 10.

Five types of attributes: including global and local ones, were investigated to evaluate the performance of our method. The Cycle-GAN[5], which consists of two generators, was reimplemented to compare with the proposed method. In the testing phase, 2000 images from the attribute-positive class and attribute-negative class were selected, each category with 1000 images. Noted that all these selected images for testing were unseen during the training phase.

*3.2. Attribute Inverse*

In the experiments, we first evaluated the proposed method by visual quality. Three types of local attributes, *eyeglasses*, *mouth_open* and *no_beard* and two types of global attributes, *young* and *male* were selected for qualitative analysis. The overall observations of the modified results are showed in figure 2. As shown in figure 2, the results by our method preserve almost all the details of original face images except the areas corresponding to the target attributes, whereas the performance of CycleGAN is not so desirable. For the attributes *month_open* and *no_beard*, both our method and CycleGAN achieve promising performance. However, for the more challenging task, manipulation on *eyeglasses*, which need to generate new object (the glasses) on the original image, our method performs much better than CycleGAN. As to *male/young* manipulation, more details such as haircut, skin texture, face shape are required to changed, and it is generally more challenging than manipulation on local attributes. In this task, the performance of our method is significant better than CycleGAN. Moreover, by employing the feature matching loss[9], the training of our method is more stable, dual learning and minimizing attribute classification loss can help the generator to attach more attention on attribute-specific areas, this is the reason why our method is well-qualified and steadier among different kinds of attributes.

*3.3. Quantitative Analysis*

To further quantify the performance of our proposed approach, we proposed a novel quantitative metric, the discrepancy of feature norm (DFN), to evaluate image quality of the generated face images. The Fréchet Inception Distance (FID) [13], a widely used metric for the evaluation of GAN-generated images, is also employed to compare with DFN.

**Table 1.** Comparisons of our method and CycleGAN on FID and DFN

| model | FID/Mean of DFN | | | | |
|---|---|---|---|---|---|
| | W/Wo beard | Mouth Open/close | W/Wo eyeglasses | Young/Old | Male/Famale |
| ours | 18.8/0.9 | 22.5/0.94 | 16.6/0.94 | 28.7/0.89 | 21.1/0.91 |
| CycleGAN | 25.0/0.86 | 28.5/0.81 | 31.6/0.83 | 38.1/0.79 | 30.7/0.86 |

*3.3.1 Evaluation of Discrepancy of Feature Norm.* We proposed the Discrepancy of Feature Norm (DFN) inspired by [14], based on the analyzation of face representations outputed from Face Recognition CNN (FRCNN). The authors in [14] explore the components of face features and present interior relationships between face image qualities and face representations using softmax loss.

Specifically, in the feature space of FRCNN, the distance of a face image to the origin is affected by a range of issues including extreme views, distortion, strong occlusion, blurring, etc. The quality of the image increases with its distance to the origin. We applied these characters to measure the quality of generated face images by calculating L2 norm of face representations captured from a state of art face recognition network. Obviously, features of high quality face images achieve higher L2 norm than those of face images with extreme defects.

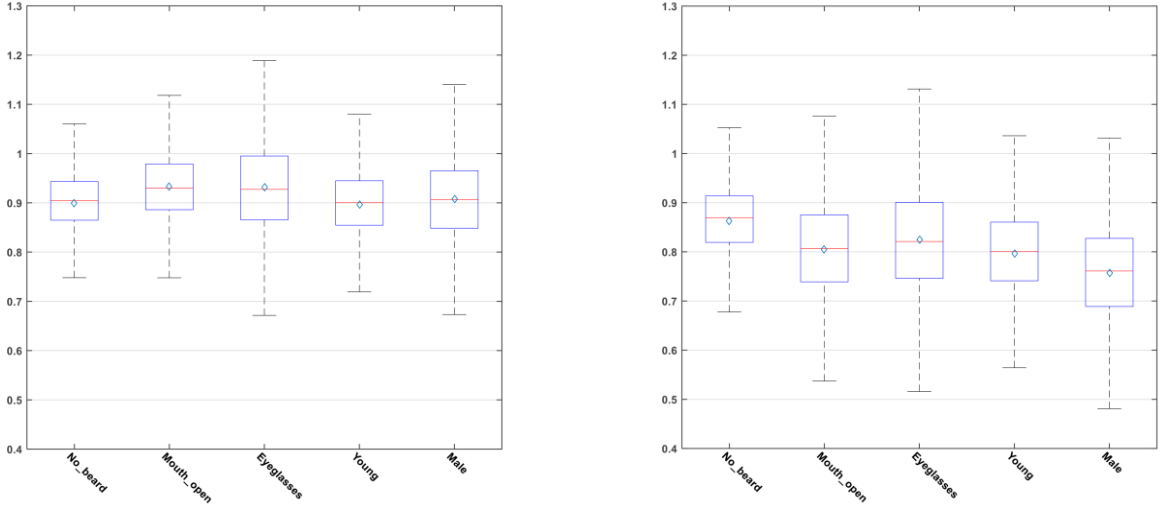

**Figure 3.** Comparisons of our method and CycleGAN on Distribution of DFN.

We extracted the face features of the testing data $x_i$ and its corresponding translated image $G(x_i)$ from the face recognition network and calculate the ratio $\gamma_i$ between the L2 norms of their face features. The ratio $\gamma_i$ is defined as

$$\gamma_i = \frac{\|\phi(G(x_i))\|_2}{\|\phi(x_i)\|_2} \tag{8}$$

where $\phi(x)$ represents the outputs of feature extraction layer, $i = 1,2,...,N$, $N$ notes the number of test data. Boxplot was used to indicate the distribution of all ratios $DFN = \{(x_0, \gamma_0),(x_1, \gamma_1),...,(x_N, \gamma_N)\}$ on the whole test dataset. It is clearly showed in figure 3 that our method obtains higher mean values than CycleGAN suggesting the generated images by our method is of better quality. For each type of attribute, our method achieves smaller interquartile range (IQR), the first quartile subtracted from the third quartile in the boxplot, which can be computed as: *IQR = 75th percentile−25th percentile* indicating the generated images by our method are much stable in quality.

*3.3.2 Evaluation of Fréchet Inception Distance.* To verify the effectiveness of DFN, we employed the Fréchet Inception Distance (FID) to further quantify the generated images and compare with DFN. The FID has been widely used for image quantitative analysis for GAN-generated images, it is defined as

$$FID(x, g) = \|\mu_x - \mu_g\|_2^2 + \mathrm{Tr}(\Sigma_x + \Sigma_g - 2(\Sigma_x \Sigma_g)^{\frac{1}{2}}) \tag{9}$$

where $(\mu_g, \Sigma_g)$ and $(\mu_x, \Sigma_x)$ are the covariance and mean of the sample representations from model distribution and data distribution, respectively. FID is used to evaluate the discrepancies between the original real images and its generated attribute-inverted images. We extracted the 2048-dimensional activations of the pool3 layer in Inception-v3 and leveraged Principal Component Analysis (PCA)[15] to output 1024-dimentional features. These features are used to calculate FID. The FID results, along with DFN results are showed in table 1, where we can observe the FID scores achieved by our method are much lower than those of CycleGAN, meaning the image qualities generated by our method are better than CycleGAN. Note these results are highly consistent with the results of the quantitative metric we proposed, which prove the proposed DFN is also suitable for image quality evaluation.

## 4. Conclusion

We have proposed a novel approach for image-to-image translation between two face attribute domains using one single generator without conditional inputs. We have analyzed the information shared in the U-net and modified it for better application. The multi-task of discriminator on attribute classification and forward-backward consistency strategy allow the generator to only focus on attribute relevant areas while keep other regions untouched. Experiments results from visual quality evaluation and our proposed quantitative evaluation metric have showed that our method can successfully capture the "mode" of the input image and transform it to its inverse "mode".

## 5. References

[1] Gatys, L.A., Ecker, A.S. and Bethge, M. 2016. Image Style Transfer Using Convolutional Neural Networks. *2016 IEEE Conference on Computer Vision and Pattern Recognition (CVPR)*, 2414-2423.
[2] Goodfellow, I.J., Pouget-Abadie, J., Mirza, M., Xu, B., Warde-Farley, D., Ozair, S., Courville, A.C. and Bengio, Y. 2014. Generative Adversarial Nets. *NIPS*.
[3] Perarnau, G., Weijer, J.V., Raducanu, B. and Álvarez, J.M. 2016. Invertible Conditional GANs for image editing. *ArXiv, abs/1611.06355*.
[4] Isola, P., Zhu, J., Zhou, T. and Efros, A.A. 2016. Image-to-Image Translation with Conditional Adversarial Networks. *2017 IEEE Conference on Computer Vision and Pattern Recognition (CVPR)*, 5967-5976.
[5] Zhu, J., Park, T., Isola, P. and Efros, A.A. 2017. Unpaired Image-to-Image Translation Using Cycle-Consistent Adversarial Networks. *2017 IEEE International Conference on Computer Vision (ICCV)*, 2242-2251.
[6] Odena, A., Olah, C. and Shlens, J. 2016. Conditional Image Synthesis with Auxiliary Classifier GANs. *ICML*.
[7] Zhang, L., Ji, Y. and Lin, X. 2017. Style Transfer for Anime Sketches with Enhanced Residual U-net and Auxiliary Classifier GAN. *2017 4th IAPR Asian Conference on Pattern Recognition (ACPR)*, 506-511.
[8] Choi, Y., Choi, M., Kim, M., Ha, J., Kim, S. and Choo, J. 2017. StarGAN: Unified Generative Adversarial Networks for Multi-domain Image-to-Image Translation. *2018 IEEE/CVF Conference on Computer Vision and Pattern Recognition*, 8789-8797.
[9] Wang, T., Liu, M., Zhu, J., Tao, A., Kautz, J. and Catanzaro, B. 2017. High-Resolution Image Synthesis and Semantic Manipulation with Conditional GANs. *2018 IEEE/CVF Conference on Computer Vision and Pattern Recognition*, 8798-8807.
[10] Zeiler, M.D. and Fergus, R. 2013. Visualizing and Understanding Convolutional Networks. *ArXiv, abs/1311.2901*.
[11] Liu, Z., Luo, P., Wang, X. and Tang, X. 2014. Deep Learning Face Attributes in the Wild. *2015 IEEE International Conference on Computer Vision (ICCV)*, 3730-3738.
[12] Masko, D. and Hensman, P. 2015. The Impact of Imbalanced Training Data for Convolutional Neural Networks.
[13] Heusel, M., Ramsauer, H., Unterthiner, T., Nessler, B. and Hochreiter, S. 2017. GANs Trained by a Two Time-Scale Update Rule Converge to a Local Nash Equilibrium. *NIPS*.
[14] Parde, C.J., Castillo, C.D., Hill, M.Q., Colon, Y.I., Sankaranarayanan, S., Chen, J. and O'Toole, A.J. 2016. Deep Convolutional Neural Network Features and the Original Image. *ArXiv, abs/1611.01751*.
[15] Smith, L. 2002. A tutorial on Principal Components Analysis.
[16] Tu X, Zhao J, Jiang Z, et al. Joint 3D Face Reconstruction and Dense Face Alignment from A Single Image with 2D-Assisted Self-Supervised Learning[J]. arXiv preprint arXiv:1903.09359, 2019.
[17] Tu X, Zhang H, Xie M, et al. Deep Transfer Across Domains for Face Anti-spoofing[J]. arXiv preprint arXiv:1901.05633, 2019.

[18] Tu X, Zhao J, Xie M, et al. Learning Generalizable and Identity-Discriminative Representations for Face Anti-Spoofing[J]. arXiv preprint arXiv:1901.05602, 2019.
[19] Tu X, Zhang H, Xie M, et al. Enhance the Motion Cues for Face Anti-Spoofing using CNN-LSTM Architecture[J]. arXiv preprint arXiv:1901.05635, 2019.
[20] Tu X, Yang F, Xie M, et al. Illumination Normalization for Face Recognition Using Energy Minimization Framework[J]. IEICE TRANSACTIONS on Information and Systems, 2017, 100(6): 1376-1379.